\documentclass[10pt, conference, compsocconf]{IEEEtran}

\IEEEoverridecommandlockouts                              

\usepackage{times}
\usepackage{epsfig}
\usepackage{epstopdf}
\usepackage{graphicx}
\usepackage{amsmath}
\usepackage{amssymb}
\usepackage{multirow}
\DeclareGraphicsExtensions{.eps}

\usepackage{subcaption}

\usepackage[pagebackref=true,breaklinks=true,letterpaper=true,colorlinks=true,bookmarks=true]{hyperref}

\title{Modular Decomposition and Analysis of Registration based Trackers}

\author{ Abhineet Singh, Ankush Roy, Xi Zhang, Martin Jagersand\\
Department of Computing Science\\
University of Alberta, Edmonton, Canada\\
{\tt\small \{asingh1, ankush2, xzhang6\}@ualberta.ca, jag@cs.ualberta.ca}
}

\begin{document}

\maketitle

\begin{abstract}

This paper presents a new way to study registration based trackers by decomposing them into three constituent sub modules:
appearance model, state space model and search method. 
It is often the case that when a new tracker is introduced in literature, it only contributes to one or two of these sub modules while using existing methods for the rest. Since these are often selected arbitrarily by the authors, they may not be optimal for the new method. In such cases, our breakdown can help to experimentally find the best combination of methods for these sub modules while also providing a framework within which the contributions of the new tracker can be clearly demarcated and thus studied better. 
We show how existing trackers can be broken down using the suggested methodology and compare the performance of the default configuration chosen by the authors against other possible combinations to demonstrate the new insights that can be gained by such an approach. We also present an open source system that provides a convenient interface to plug in a new method for any sub module and test it against all possible combinations of methods for the other two sub modules while also serving as a fast and efficient solution for practical tracking requirements.
\end{abstract}


\section{Introduction}
\label{introduction}


Since its inception, research in object tracking has focused on presenting new tracking algorithms to address specific challenges in a wide variety of application domains like surveillance, targeting systems, augmented reality and medical analysis. However, before an algorithm can be adopted in a real life application, it needs to be extensively tested so that both its advantages and limitations can be determined. Recent studies in tracking evaluation \cite{Wu13benchmark, Kristan2015_vot15} show increasing efforts to standardize this crucial process. However, though such studies assign a global rank to each tracker, they often provide little feedback to improve these trackers since they treat them as black boxes predicting the trajectory of the object. A more useful evaluation methodology would be to have empirical validation of the tracker's design or point out its shortcomings. 

An exhaustive analysis of learning based trackers is admittedly a daunting and impracticable task as these often use widely varying techniques that have little in common. This, however, is not true for registration based trackers \cite{Lucas81lucasKanade, Baker04lucasKanade_paper} which - as we show in this work - can be decomposed into three well defined modules, thus making their systematic analysis feasible. These trackers are generally faster and more precise than learning based trackers \cite{Roy2015_tmt} which makes them more suitable for applications such as robotic manipulations, visual servoing and SLAM, where multiple trackers are used in parallel. On the other hand, lacking an online learning component, they are known to be non robust to changes in the object's appearance and prone to failure in the presence of motion blur, occlusion, lighting variations or viewpoint changes. As a result, they are less popular in the vision community and often underrepresented in the aforementioned studies, thus making such an evaluation particularly useful for applications where learning based trackers are unsuitable.
A detailed analysis, with a test framework in registration based tracking, to the best of our knowledge. has never been attempted before.

Many reported studies in this domain \cite{Lucas81lucasKanade, Baker04lucasKanade_paper, Benhimane07_esm_journal} have introduced new methods for only one of the three submodules without exploring the full extent of their contributions. For instance, Baker et. al \cite{Baker04lucasKanade_paper} reported a compositional update scheme for the state parameters $\mathbf{p}$ (Eq. \ref{reg_tracking}) instead of the additive scheme used in \cite{Hager98parametricModels}, but never experimented with different AMs.
Conversely, Richa et. al \cite{Richa11_scv_original} showed an improvement over the existing efficient second order minimization \cite{Benhimane07_esm_journal} approach by using the sum of conditional variance as the similarity metric instead of the sum of squared differences. Similarly, Dame et. al \cite{Dame10_mi_ict} used mutual information while Scandaroli et. al \cite{Briechle2001_ncc_template_matching} used normalized cross correlation with the inverse compositional method of \cite{Baker04lucasKanade_paper}. However, neither of them tested their similarity measures with other search methods even though the latter had previously been shown to be a good metric when used with the standard Lucas Kanade type tracker \cite{Briechle2001_ncc_template_matching}.  

Finding the optimal combination of methods for any tracking algorithm is a two step process. First, the sub module where the algorithm's main contribution lies needs to be determined, using, for instance, the method employed in \cite{Wu13benchmark}. Second, all possible combinations for the other sub modules that are compatible with this algorithm (since not all methods for different sub modules work with each other) need to be enumerated and evaluated. A generic framework would thus be useful to avoid such fragmentation. 

To summarize, following are the main contributions of this work:
\begin{itemize}
\item Empirically test different combinations of submodules leading to several interesting observations and insights that were missing in the original papers. Experiments are done using two large datasets with over 77,000 frames in all to ensure their statistical significance.

\item Report for the first time, to the best of our knowledge, results comparing robust similarity metrics \cite{Richa12_robust_similarity_measures}, with traditional SSD type measures.

\item Compare formulations against popular online learning based trackers to validate their usability in precise tracking applications.

\item Provide an open source tracking framework \footnote{available online at \url{http://webdocs.cs.ualberta.ca/~vis/mtf/}} using which all results can be reproduced and which, owing to its efficient C++ implementation, can also be used to address practical tracking requirements.
\end{itemize}


\section{Descriptions of submodules}
\begin{figure}[h]
	\begin{center}
	\includegraphics[width=2.8in]{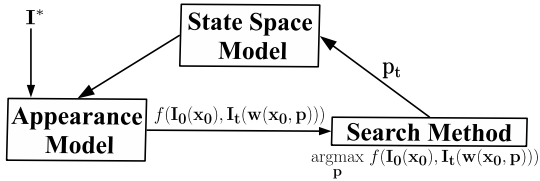}
	\caption{Modular breakdown of a registration based tracker assuming there is no dynamic update to the appearance model. This shows how different components work, as formulated in Eq \ref{reg_tracking}}
	\label{flow}
	\end{center}
\end{figure}

A registration based tracker can be decomposed into three sub modules: appearance model (\textbf{AM}),   state space model (\textbf{SSM}) and search method (\textbf{SM}). Figure \ref{flow} shows how these modules work together in a complete tracking system assuming there is no model update wherein the appearance model of the object is updated as tracking progresses.  

When a geometric transform  $\mathbf{w}$ with parameters $\mathbf{p}=(p_1, p_2, ..., p_S)$ is applied to an image patch $\mathbf{x}$, the transformed patch is denoted by $\mathbf{x}'=\mathbf{w}(\mathbf{x}, \mathbf{p})$ and the corresponding pixel values in image $ I $ as $\mathbf{I}(\mathbf{w}(\mathbf{x}, \mathbf{p}))$. Tracking can then be formulated (Eq \ref{reg_tracking}) as a search problem where we need to find the optimal transform parameters $\mathbf{p_t}$ for an image $I_t$ that maximize the similarity, measured by a suitable metric $f$, between the target patch $\mathbf{I^*} = \mathbf{I_0}(\mathbf{w}(\mathbf{x},\mathbf{p_0}))$ and the warped image patch $\mathbf{I_t}(\mathbf{w}(\mathbf{x},\mathbf{p_t}))$. 
\begin{equation}
\begin{aligned}
\label{reg_tracking}
\mathbf{p_t} = \underset{\mathbf{p}} {\mathrm{argmax}} ~f(\mathbf{I^*},\mathbf{I_t}(\mathbf{w}(\mathbf{x},\mathbf{p})))
\end{aligned}
\end{equation}
We refer to the similarity metric $ f $, the warp function $ \mathbf{w} $ and the algorithm that maximizes Eq \ref{reg_tracking} respectively as AM, SSM and SM.
A more detailed description of these submodules follows.

\subsection{Search Method}
\label{searchMethod}
This is the optimization procedure that searches for the warped patch in the the current image that best matches the original template. Gradient descent is the most popular optimization approach used in tracking due to its speed and simplicity and is the basis of the classic Lucas Kanade (LK) tracker \cite{Lucas81lucasKanade}. This algorithm can be formulated in four different ways \cite{Baker04lucasKanade_paper} depending on which image is searched for the warped patch - $ I_t $ or $ I_0 $ - and how the parameters of the warping function are updated in each iteration - additive or compositional. The four resulting variants - forward additive (\textbf{FALK}) \cite{Lucas81lucasKanade}, inverse additive (\textbf{IALK}) \cite{Hager98parametricModels},  forward compositional (\textbf{FCLK}) \cite{Szeliski06_fclk_extended} and inverse compositional (\textbf{ICLK}) \cite{Baker04lucasKanade_paper} - were  analyzed mathematically and shown to be equivalent to first order terms in \cite{Baker04lucasKanade_paper}. Here, however, we show experimental results proving that their performance on real video benchmarks is quite different (Sec. \ref{res_sm}). 

A relatively recent update to this approach was in the form of Efficient Second order Minimization (\textbf{ESM}) \cite{Benhimane07_esm_journal} technique that tries to make the best of both inverse and forward formulations by using the mean of the initial and current Jacobians. 
We would like to mention here that, even though the authors of \cite{Benhimane07_esm_journal} used $\mathcal{SL}3$ parameterization for their ESM formulation and gave theoretical proofs as to why it is essential for this SM, we have used standard parameterization (i.e. using matrix entries \cite{Szeliski06_fclk_extended, Baker04lucasKanade_paper}) for all our experiments since, as we show later (Sec. \ref{res_ssm}), ESM actually performs identically with several different parameterizations.

Further, since the standard formulations for these SMs using the Gauss Newton Hessian \cite{Lucas81lucasKanade, Baker04lucasKanade_paper, Benhimane07_esm_journal} do not work with any AMs besides SSD \cite{Dame10_mi_ict,Scandaroli2012_ncc_tracking}, a modified version with the so called \textit{Hessian after convergence} \cite{Dame10_mi_ict,Scandaroli2012_ncc_tracking} has been used for our experiments. Also, the extended formulation for ESM reported in \cite{Brooks10_esm_ic,Scandaroli2012_ncc_tracking} has been used instead of the original one in \cite{Benhimane07_esm_journal}. The exact formulations used can be found in \cite{singh16_mtf}.

Nearest neighbor search (NN) is another SM that has recently been used for tracking \cite{Dick13nn} thanks to the FLANN library \cite{Muja2009_flann} that makes real time search feasible. Since the performance of stochastic SMs like NN depends largely on the number of random samples used, we have reported results with 1000 and 10000 samples, with the respective SMs named as \textbf{NN1K} and \textbf{NN10K}.  Further, this method tends to give jittery and unstable results when used by itself due to the very limited search space and so was used in conjunction with a gradient descent type SM in \cite{Dick13nn} to create a composite tracker that performs better than either of its constituents. As in \cite{Dick13nn}, we have used ICLK as this second tracker due to its speed and the resultant composite SM is named \textbf{NNIC}. Unlike NN, results for NNIC are only reported using 1000 samples for NN as NN10K is too slow to be combined with ICLK.


\subsection{Appearance Model}
\label{appearanceModel}

This is the similarity metric defined by the function $f$ in Eq. \ref{reg_tracking} using which the SM compares different warped patches from the current image to get the closest match with the original template. 

The sum of squared differences (\textbf{SSD}) \cite{Lucas81lucasKanade,Baker04lucasKanade_paper,Benhimane07_esm_journal} or the L2 norm of pixel differences is the AM used most often used in literature especially with SMs based on gradient descent search due to its simplicity and the ease of computing its derivatives. However, the same simplicity also makes it vulnerable to providing false matches when the object's appearance changes due to factors like lighting variations, motion blur and occlusion. 

To address these issues, more robust AMs have been proposed including Sum of Conditional Variance (\textbf{SCV}) \cite{Richa11_scv_original}, Normalized Cross Correlation (\textbf{NCC}) \cite{Scandaroli2012_ncc_tracking}, Mutual Information (\textbf{MI}) \cite{Dowson08_mi_ict,Dame10_mi_ict} and Cross Cumulative Residual Entropy (\textbf{CCRE}) \cite{Wang2007_ccre_registration,Richa12_robust_similarity_measures}, all of which supposedly provide a degree of invariance to changes in illumination. There also exists a slightly different formulation of the former known as Reversed SCV (\textbf{RSCV}) \cite{Dick13nn} where $ \mathbf{I_t} $ is updated rather than $ \mathbf{I_0} $. There has also been a recent extension to it called \textbf{LSCV} \cite{Richa14_scv_constraints} that uses multiple joint histograms from corresponding sub regions within the target patch to achieve greater robustness to localized intensity changes. 
It has further been shown \cite{Ruthotto2010_thes_ncc_equivalence} that maximizing NCC between two images is equivalent to minimizing the SSD between two z-score \cite{Jain2005_ncc_z_score} normalized images. We consider the resultant formulation as a different AM called Zero Mean NCC (\textbf{ZNCC}). 

It may ne noted that these AMs can be divided into 2 distinct categories - those that use some form of the L2 norm as the similarity function - SSD, SCV, RSCV, LSCV and ZNCC - and those that do not - MI, CCRE and NCC. The latter are henceforth called robust models after \cite{Richa12_robust_similarity_measures}.


\subsection{State Space Model}
\label{stateSpace}
The SSM represents the set of allowable image motions of the tracked object and thus embodies any constraints that are placed on the search space of warp parameters to make the optimization more efficient. This includes both the degrees of freedom (DOF) of allowed motion, as well as the actual parameterization of the warping function. For instance the ESM tracker, as presented in \cite{Benhimane07_esm_journal}, can be considered to have a different SSM than conventional LK type trackers \cite{Lucas81lucasKanade,Baker04lucasKanade_paper} even though both involve 8 DOF homography, since it uses the $\mathcal{SL}3$ parameterization rather than the actual entries of the corresponding matrix.
We model 7 different SSMs - translation, isometry, similitude, affine and homography \cite{Szeliski06_fclk_extended} along with two extra parameterizations of homography - $ \mathcal{SL}3 $ and corner based (using x,y coordinates of the four corners of the bounding box).

The advantage of using higher DOF SSM is achieving greater precision in the aligned warp since transforms that are higher up in the hierarchy \cite{Hartley04MVGCV} can better approximate the projective transformation process that captures the relative motion between the camera and the object in the 3D world into the 2D images. 
However, there are two issues with having to estimate more parameters - the iterative search takes longer to converge making the tracker slower and the search process becomes more likely to either diverge or end up in a local optimum causing the tracker to be less stable and more likely to lose track. The latter is a well known phenomenon with LK type trackers \cite{Bouguet01} whose higher DOF variants are usually less robust.

It may be noted that this sub module differs from the other two in that it does not admit new methods in the conventional sense and may even be viewed as a part of the SM with the two being often closely intertwined in practical implementations.
However, though the SSMs used in this work are limited to the standard hierarchy of geometric transformations, more complex models like piecewise projective transforms do exist and it is also theoretically possible to impose novel constraints on the search space that can significantly decrease the search time while still producing sufficiently accurate results. The fact that such a constraint will be an important contribution in its own right justifies the use of SSM as a sub module in this work to motivate further research in this direction.



\section{Experimental Results and Analysis}
\label{experiments}
\subsection{Dataset and Error Metric}
Two publicly available datasets have been used to analyze the trackers:
\begin{enumerate}
\item Tracking for Manipulation Tasks (\textbf{TMT}) dataset \cite{Roy2015_tmt} that contains videos of some common tasks performed at several speeds and under varying lighting conditions. It has 109 sequences with a total of 70592 frames. 
\item  Visual Tracking Dataset provided by \textbf{UCSB} \cite{Gauglitz2011_ucsb} that has 96 short sequences of different challenges in object tracking with a total of 6889 frames. The sequences here are more challenging but also rather artificial since they were created specifically to address various challenges rather than represent realistic scenarios.
\end{enumerate} 
Both these datasets have full pose (8 DOF) ground truth data which makes them suitable for evaluating high precision trackers that are the subject of this study. In addition, we use \textbf{Alignment Error} ($E_{AL}$) \cite{Dick13nn} as metric to compare tracking result with the ground truth since it accounts for fine misalignments of pose better than other common measures like center location error and Jaccard index.

\subsection{Evaluation Measure}
We measure a tracker's overall accuracy through its \textbf{success rate} (SR) which is defined as the fraction of total frames where the tracking error $E_{AL}$ is less than a threshold of $t_p$ pixels. Formally, 
$ SR = \frac{|S|}{|F|} $
where $ S = \{f^{i} \in F : E_{AL}^{i} < t_p  \}$, $F$ is the set of all frames and $E_{AL}^{i}$ is the error in the $i^{th}$ frame $f^{i}$. Since we have far too many sequences to present results for each, we instead report an overall summary of performance by averaging the success rates over all the sequences in both datasets, i.e. $ F $ is treated as the set of all frames in TMT and UCSB with $ |F| = 70592 + 6889 - 205 = 77276 $ - we do not consider the first frame in each sequence, where the tracker is initialized, for computing the SR. Finally, we evaluate SR for several values of $ t_p $ ranging from 0 to 20 and study the resulting SR vs. $ t_p $ plot to get an overall idea of how precise and robust a tracker is.
 
\subsection{Parameters Used}
All results have been generated using a fixed sampling resolution of $ 50{\times}50 $ irrespective of the tracked object's size. The input images were smoothed using a Gaussian filter with a $ 5{\times}5 $ kernel before being fed to the trackers. Iterative SMs were allowed to perform a maximum of $ 30 $ iterations per frame but only as long as the L2 norm of the change in bounding box corners in each iteration remained greater than $ 0.001 $. For the NN tracker, a standard deviation of $ 0.05 $ was used for generating the random warps. The learning based trackers whose results are reported in Sec. \ref{res_ssm} were run using default settings provided by their respective authors. All speed tests were run on a 2.66 GHz Intel Core 2 Quad Q9450 machine with 4 GB of RAM. No multi threading was used.

\subsection{Results}
The results presented in this section are organized into three sections corresponding to the three sub modules. In each of these, we present and analyze results comparing different methods for implementing the respective sub module with one or more combinations of methods for the other sub modules. SSM is fixed to homography for the first two sections. 
\label{results}
\subsubsection{Search Methods}
\label{res_sm}
\begin{figure}[h]
\centering
\includegraphics[width=\linewidth]{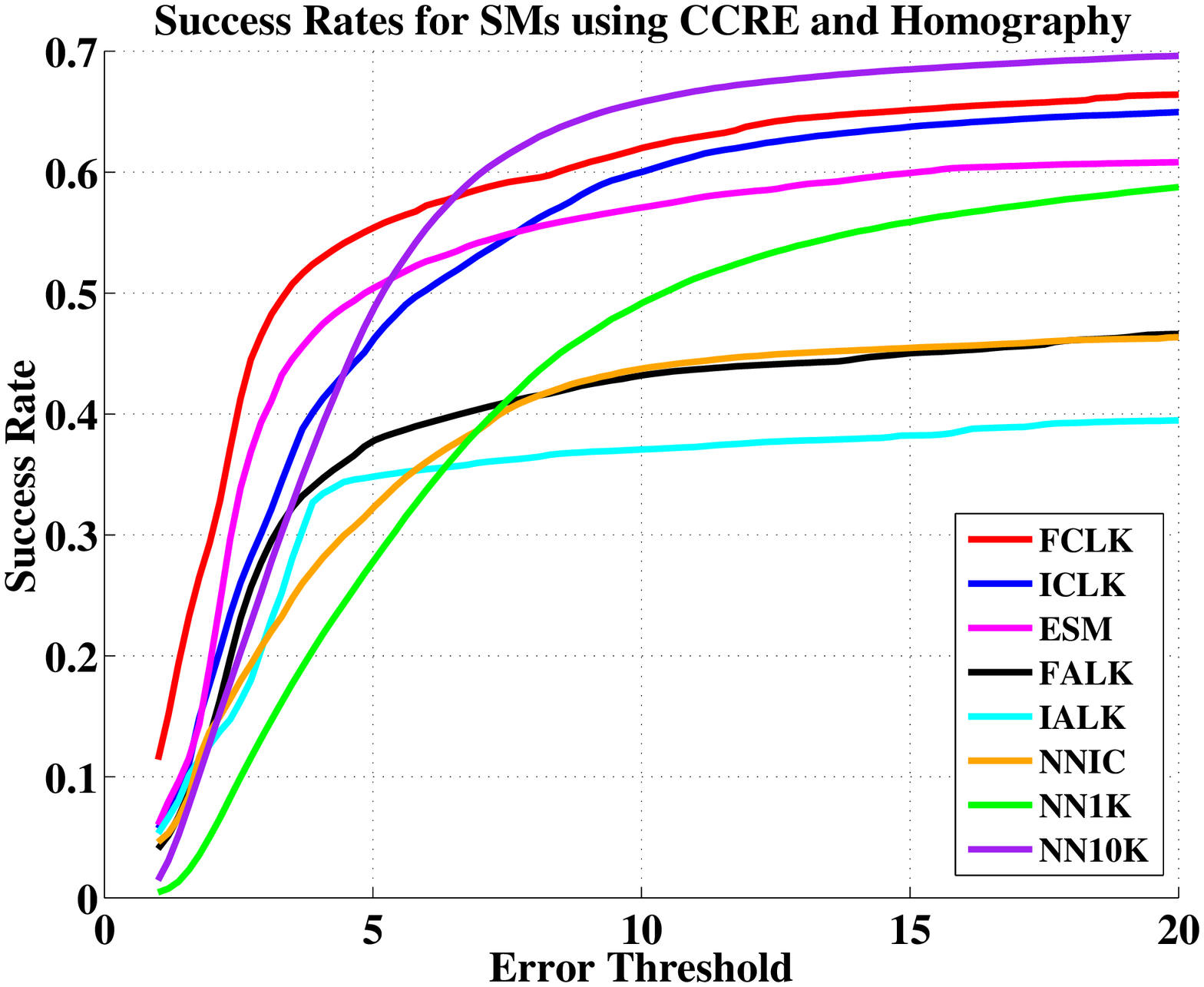}
\caption{Success rates for SMs using CCRE with Homography}
\label{fig_sm_ccre}
\end{figure}

\begin{figure}[!htbp]
\centering
\includegraphics[width=\linewidth]{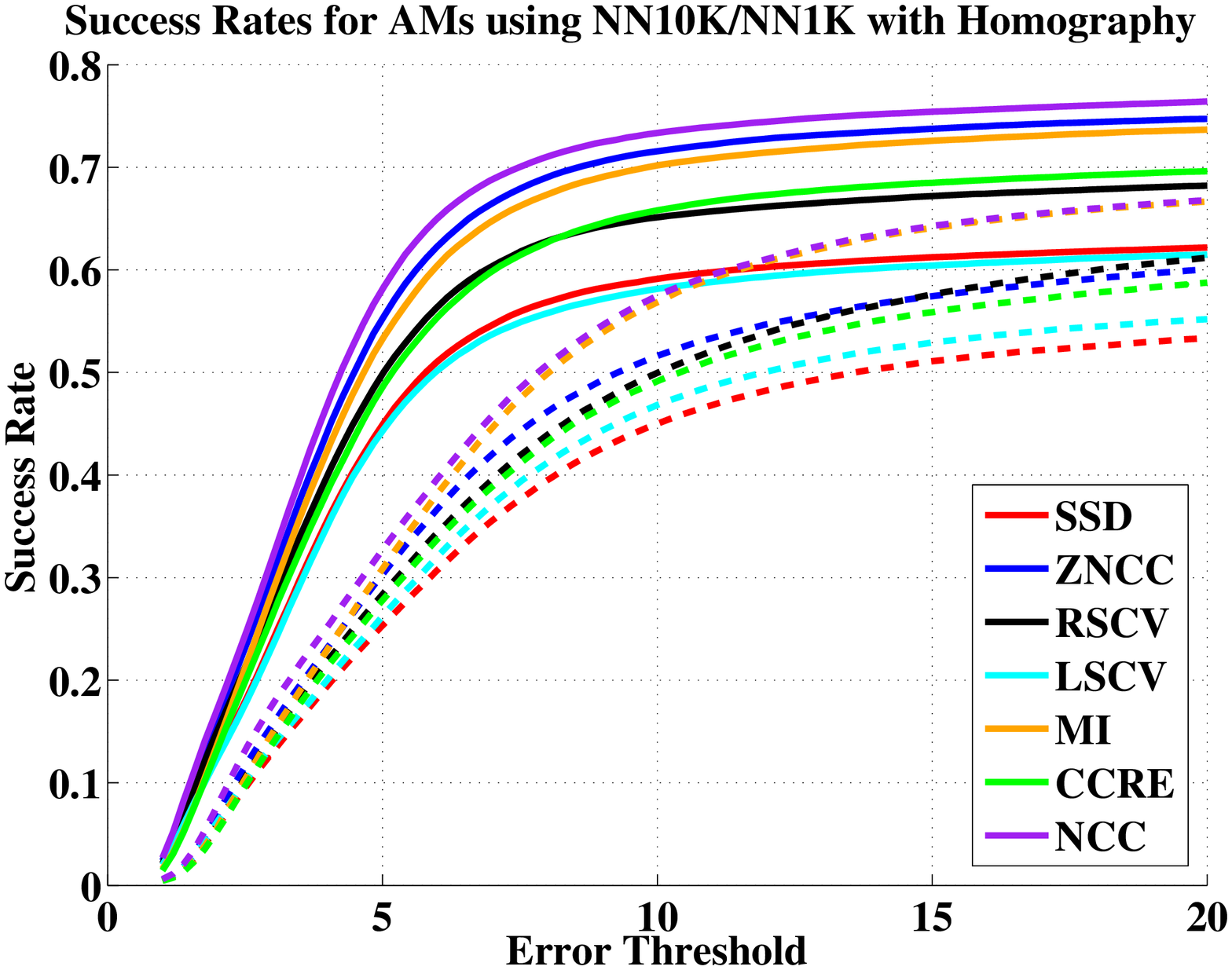}
\caption{Success rates for AMs using NN10K and NN1K with Homography represented with \textbf{solid and dashed lines} respectively. SCV, being almost identical to LSCV,  has been ommitted for clarity.}
\label{fig_am_nn}
\end{figure}

\setlength{\floatsep}{0pt}
\begin{figure*}[!htbp]
\begin{subfigure}{\textwidth}
  \centering
  \includegraphics[height=0.23\textheight]{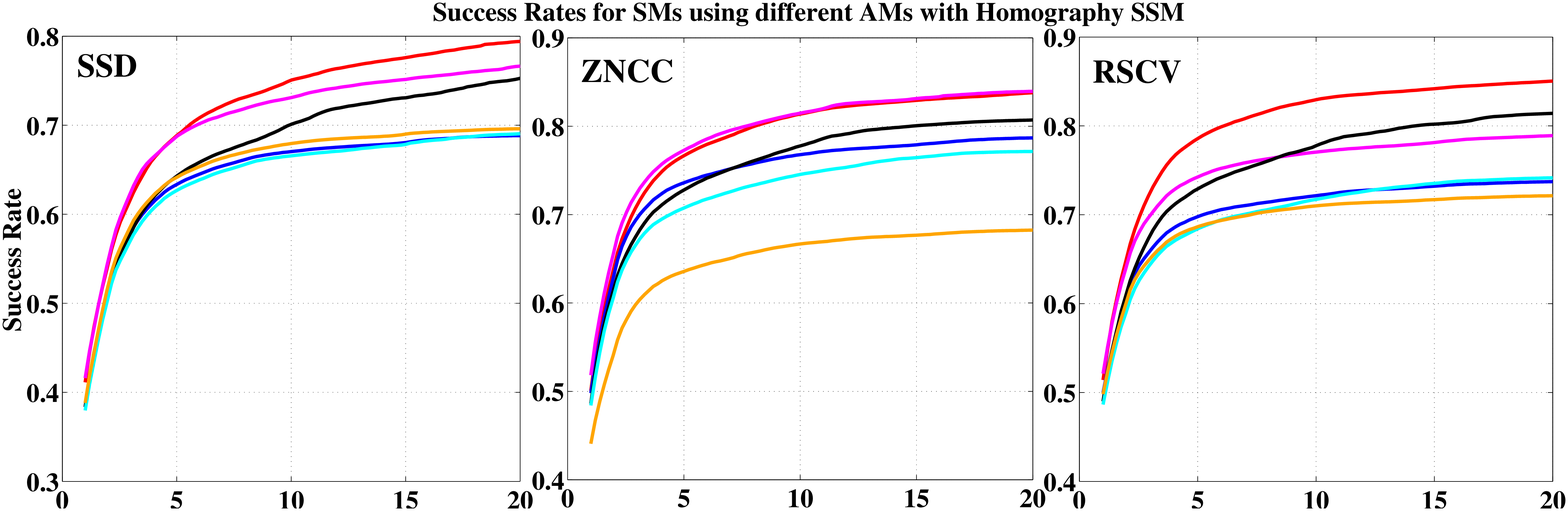}
\end{subfigure}
\begin{subfigure}{\textwidth}
  \centering
  \includegraphics[height=0.23\textheight]{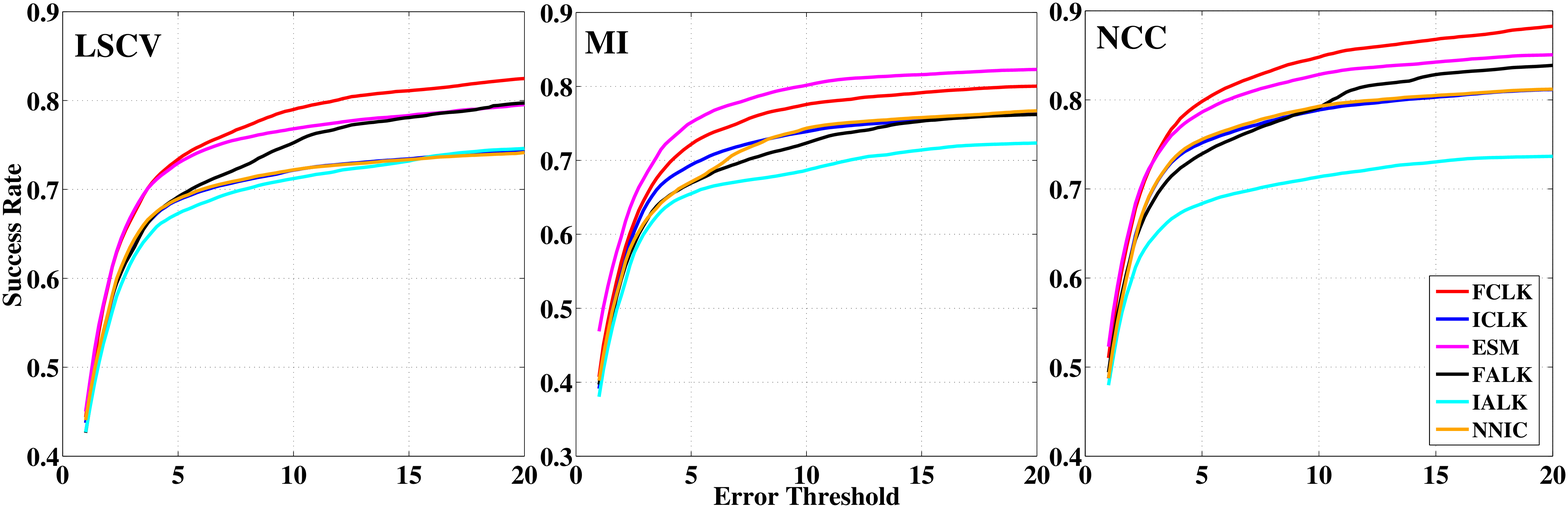}
\end{subfigure}
\caption{Success rates for SMs using Homography SSM and different AMs. Best viewed on a high resolution screen.}
\label{fig_sm}
\end{figure*}
\begin{figure*}[!htbp]
\begin{subfigure}{\textwidth}
  \centering
  \includegraphics[height=0.23\textheight]{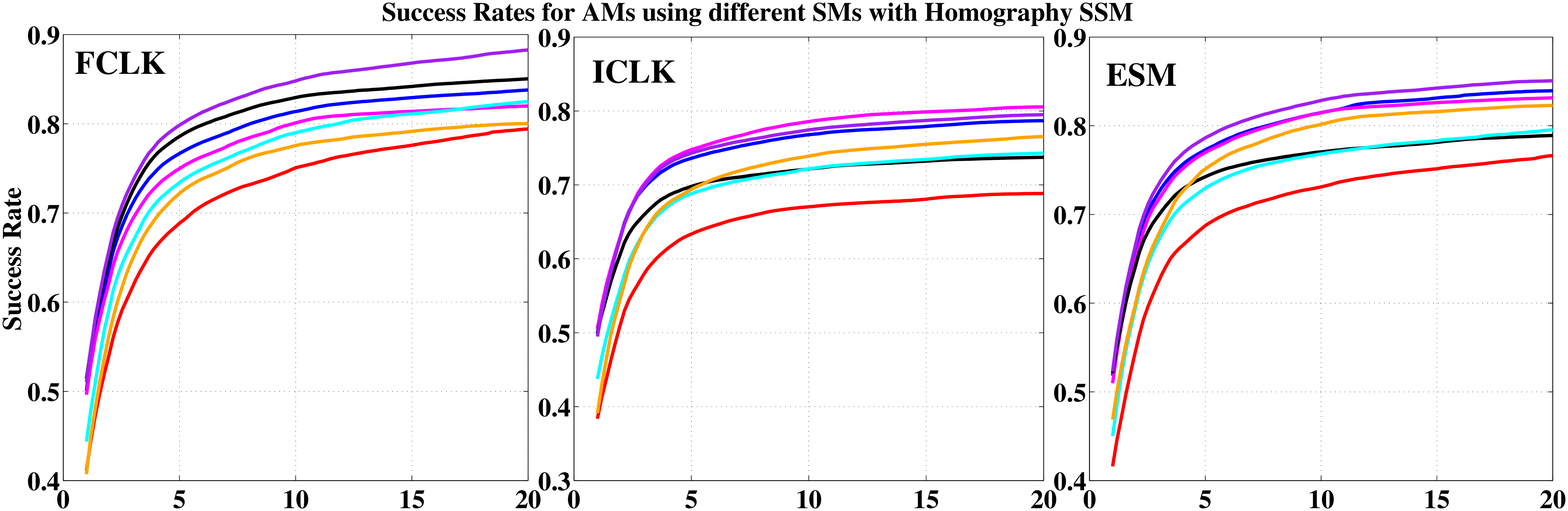}
\end{subfigure}
\begin{subfigure}{\textwidth}
  \centering
  \includegraphics[height=0.23\textheight]{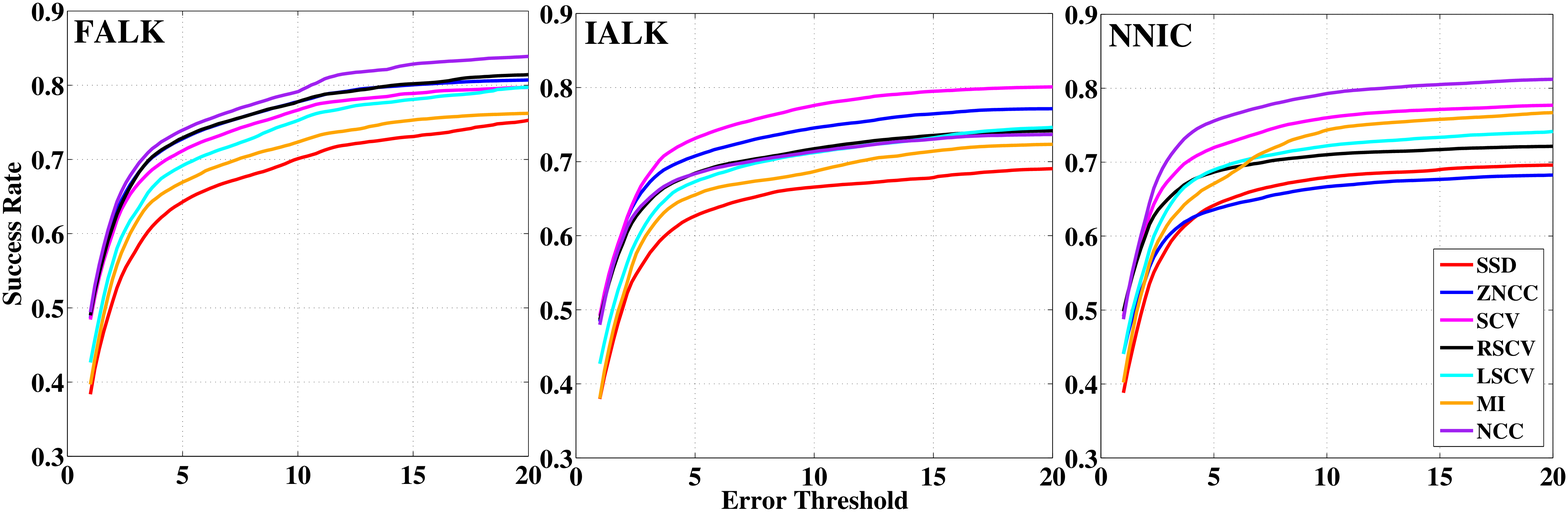}
\end{subfigure}
\caption{Success rates for AMs using Homography SSM and different SMs. Best viewed on a high resolution screen.}
\label{fig_am}
\end{figure*}
Fig. \ref{fig_sm} presents the results for all SMs except NN1K and NN10K which are presented separately in Fig. \ref{fig_am_nn}. This separation was needed because NN, due to its stochastic nature, tends to have significantly lower SR for smaller thresholds than other SMs. In order to maximize the visibility of individual curves in the various plots within Fig. \ref{fig_sm}, the y axis in each has been limited to the range where the curves in that plot actually lie. Inclusion of NN results here would have caused this range to increase significantly, thus decreasing the separation between these curves and making analysis more difficult. SCV and CCRE results are excluded here too, the former because they are very similar to LSCV while the latter are presented separately in Fig. \ref{fig_sm_ccre} for the same reason as NN but now pertaining to Fig. \ref{fig_am}.

Several interesting observations can be made from Figs. \ref{fig_sm} and \ref{fig_sm_ccre}. Firstly, we see that the four variants of LK do not perform identically - FCLK is the best for all AMs and is significantly better than FALK especially for smaller thresholds. ICLK with IALK, on the other hand, are more contentious, being very similar for three AMs - SSD, RSCV and LSCV - but ICLK being appreciably better for the other four. This is especially true for CCRE where it is almost equivalent to FCLK for larger $ t_p $ and much better than both the additive variants. This finding contradicts the equivalence between these variants that was reported in \cite{Baker04lucasKanade_paper} and justified there using both theoretical analysis and experimental results. The latter, however, were only performed on synthetic images and even the former used several approximations. So, it is perhaps not surprising that this supposed equivalence does not hold under real world conditions. 

Secondly, we note that ESM fails to outperform FCLK for any AM except MI and even there it does not lead by much. This fact too emerges in contradiction to the theoretical analysis in \cite{Benhimane07_esm_journal} where ESM was shown to have second order convergence and so should be better than first order methods including FCLK. It might be argued that the extended version of ESM \cite{Brooks10_esm_ic,Scandaroli2012_ncc_tracking} used here might not possess the characteristics of the formulation described in \cite{Benhimane07_esm_journal} but we conducted extensive experiments with that exact formulation too and can confirm that the version reported here performs identically to that one.

Thirdly, we see that NNIC does not perform better than ICLK on any of the AMs and is in fact significantly poorer with ZNCC. This yet again does not agree with the results reported in \cite{Dick13nn} using both static experiments and the Metaio dataset \cite{Lieberknecht2009_metaio}. We have already seen in our first observation that static experiments may not always agree with real world tests and it must be admitted that sequences in the Metaio benchmark are highly artificial in nature as they neither represent real tasks nor include an actual background around the tracked patch. We did try to perform experiments on this dataset to check for possible bugs in our implementation but unfortunately the Metaio evaluation service is no longer available. However, to the best of our belief, there is no such bug and the discrepancy does indeed arise from the differences between artificial and real world benchmarks.

Fourthly, we can note that both additive LK variants and especially IALK perform much poorer compared to the compositional variants with the robust AMs than with the SSD like AMs. This is probably to be expected since the Hessian after convergence approach used for extending the Gauss Newton method to these AMs does not make as much sense for additive formulations \cite{Dame10_mi_thesis}.

We conclude this section by examining the effect of number of samples on NN as well as its relative performance to gradient descent SMs from Figs. \ref{fig_sm_ccre} and \ref{fig_am_nn}. We can see by comparing these plots to Fig. \ref{fig_am} that NN performs better relative to the latter with the robust AMs and in fact CCRE actually fares best with NN10K for larger $ t_p $. This might indicate that the poor performance of CCRE, and to an extent MI, with LK type trackers has more to do with gradient descent optimization itself rather than some limitation of these AMs as good similarity metrics.
The gain in performance between NN10K and NN1K though  seems to be similar for all AMs as it is caused by an improved coverage of the SSM search space and so should depend only on that.

\subsubsection{Appearance Models}
\label{res_am}
Fig. \ref{fig_am} shows the SR curves for all AMs except CCRE whose results are in Fig. \ref{fig_sm_ccre} for reasons already mentioned in the previous section.  This reason itself is the most obvious point to be noted by comparing Figs. \ref{fig_am} and \ref{fig_sm_ccre} - that CCRE, even though it is the most sophisticated and computationally expensive AM, performs much poorer than other AMs with all SMs except those based on NN. Another interesting fact is that it actually performs far worse with NNIC than it does with either NN1K or ICLK which is very unexpected as the composite tracker uses inputs from both and so should perform at least as well as the best of these. A similar phenomenon can be observed with ZNCC too. We repeated these experiments several times  but these discrepancies remained.

Further, even MI is only slightly better than SSD on average, except with NN where it is among the best, being almost at par with NCC. It is much better than CCRE, however, in spite of the two AMs differing only in the latter using a cumulative joint histogram. It seems likely that the additional complexity of CCRE along with the resultant invariance to appearance changes significantly \textit{reduces} its basin of convergence \cite{Dame10_mi_ict}. This leads to poor performance with gradient descent type SMs but, as expected, does not affect the efficacy of stochastic SMs.

The next fact to note is that NCC is the best performer with all SMs except IALK (which performs poorly with all robust AMs anyway as noted in the previous section).
We also note that, though ZNCC is supposedly equivalent to NCC \cite{Ruthotto2010_thes_ncc_equivalence} and also has a wider basin of convergence due to its SSD like formulation, it usually does \textit{not} perform as well as NCC. 
However both ZNCC and NCC are almost always better than SCV and its extensions LSCV/RSCV.

This last observation is rather contrary to expectations since SCV is supposedly more robust against lighting changes due to its use of joint probability distributions while ZNCC is merely the L2 norm between the pixel values normalized to have zero mean and unit variance. We can note too that LSCV, notwithstanding, its reported \cite{Richa14_scv_constraints} increased invariance to localized intensity changes, fails to offer any improvement over either SCV or RSCV even though several of the tested sequences do exhibit such lighting changes. Considering that SCV and its variants are significantly more expensive than ZNCC to compute, there seems little reason to use these instead as the computational savings from ZNCC can be used to employ other ways (i.e. higher sampling resolution or more iterations) to improve performance.

\subsubsection{State Space Models}
\label{res_ssm}
\begin{figure*}[!htbp]
\begin{center}
\includegraphics[width=\textwidth]{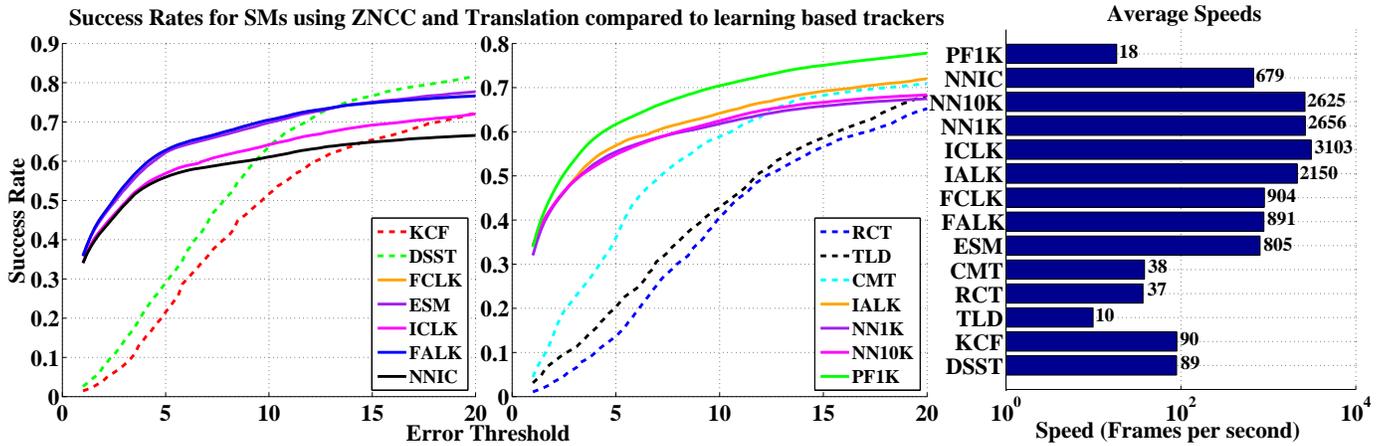}
\caption{Success Rates for SMs using ZNCC and Translation as well as for 5 learning based trackers. The former are shown with \textbf{solid} lines and the latter in \textbf{dashed} lines. 2 DOF ground truth was used for all evaluations. Note that the speed plot on the right uses \textbf{logarithmic scaling} on the x axis to increase visibility of the latter though the actual figures are mentioned too.}
\label{fig_sm_rscv_2_gt2_speed}
\end{center}
\end{figure*}
The results presented in this section follow a slightly different format from the other two sections due to the difference in the motivations for using low DOF SSMs - the principle one being that reducing the dimensionality of the search space of warp parameters decreases the likelihood of the search process getting stuck in a local optimum, thus making the tracker more robust. The other less important motivation is that lower DOF SSMs tend to be faster since Jacobians are typically less expensive to compute.

Limiting the DOF also makes registration based trackers directly comparable to learning based trackers as these too work in low DOF search spaces. As a result, in this section, we also present results for five state of the art learning based trackers \cite{Kristan2015_vot15} - discriminative scale space tracker (\textbf{DSST}), kernelized correlation filter tracker (\textbf{KCF}), tracking-learning-detection (\textbf{TLD}), real time compressive tracker (\textbf{RCT}) and consensus-based matching of keypoints tracker (\textbf{CMT}). We have used C++ implementations of all these trackers that are fully integrated into our framework. This not only makes it easy to reproduce the results presented here and but also makes it reasonable to compare the speeds of these trackers with the faster registration based trackers since slower speed is one of the main reasons why learning trackers are often not used in robotics applications. 

Lastly, in order to make the evaluations fair, we have used \textit{lower DOF ground truths} for all accuracy results in this section. These were generated for each SSM using least squares optimization to find the warp parameters that, when applied to the initial bounding box, will produce a warped box whose alignment error ($E_{AL}$) with respect to the full 8 DOF ground truth is as small as it is possible to achieve given the constraints of that SSM. In most cases, the ground truth corners thus generated represent the best possible performance that can theoretically be achieved by any tracker that follows the constraints of that SSM. In some rare cases, however, the resulting corners can be quite unexpected so we also visually inspected all lower DOF corners and corrected any that appeared unreasonable.

Fig. \ref{fig_sm_rscv_2_gt2_speed} shows the performance of all SMs with translation SSM in terms of both accuracy, evaluated against 2 DOF ground truth, and speed, measured in terms of the average number of frames processed by the tracker per second (FPS). In addition to the SMs described in Sec. \ref{searchMethod}, results from another SM based on particle filter \cite{Isard98condensation}, generated using 1000 particles (\textbf{PF1K}), are also reported here. This is another stochastic SM like NN that, though present in our framework, only works well with translation at the time of this writing and is thus not mentioned in the previous sections. 

As expected, all the learning based trackers have low SR for smaller $ t_p $ since they are less precise in general \cite{Kristan2015_vot15}. What is more interesting, however, is that none of these trackers, with the exception of DSST, managed to surpass the best registration based trackers even for larger $ t_p $ though they did close the gap. Even DSST only managed it at the extreme tail end of the plot and by a small margin. The superiority of DSST over other learning based trackers is at least consistent with results published elsewhere \cite{Kristan2015_vot15}.

The speed comparisons in Fig. \ref{fig_sm_rscv_2_gt2_speed} clearly show the main reason why learning trackers are not suitable for high speed tracking scenarios - they are $ 10 $ to $ 30 $ times slower than their registration based counterparts. 
It is not surprising that tracking based SLAM systems like SVO \cite{Forster2014_svo} use registration based trackers as they need to track hundreds to thousands of patches per frame.
It may be noted that the speeds of the former depend on the size of the initial bounding box and so varied widely between sequences unlike the latter where a fixed sampling resolution was used. However, the mean figures reported here do provide a good idea of the general performance that can be expected from these trackers. 

Some interesting observations can  be made by comparing the different SMs too. Firstly, we see that FALK and FCLK show perfect overlap which is to be expected as the two formulations are identical for translation. Secondly, we note that NN1K and NN10K have practically identical performance in terms of both accuracy and speed. The latter is to be expected since the KD Tree index used by FLANN library \cite{Muja2009_flann} is largely independent of the number of samples - only the initialization time increases when a larger index is to be built. The former, however, is a bit more difficult to explain since NN10K does perform significantly better than NN1K with homography (Fig. \ref{fig_am_nn}). It seems, however, that more samples do not help much with low DOF search spaces as 1000 samples is already enough to cover it well and it is the \textit{quality} of samples that forms the bottleneck now.  It may be noted too that PF performs at par with the best registration trackers. This is unsurprising since PF is known to perform well with low DOF when large number of particles are available - an asset that comes at the cost of much slower speed.
\begin{figure}[h]
\begin{center}
\includegraphics[width=0.5\textwidth]{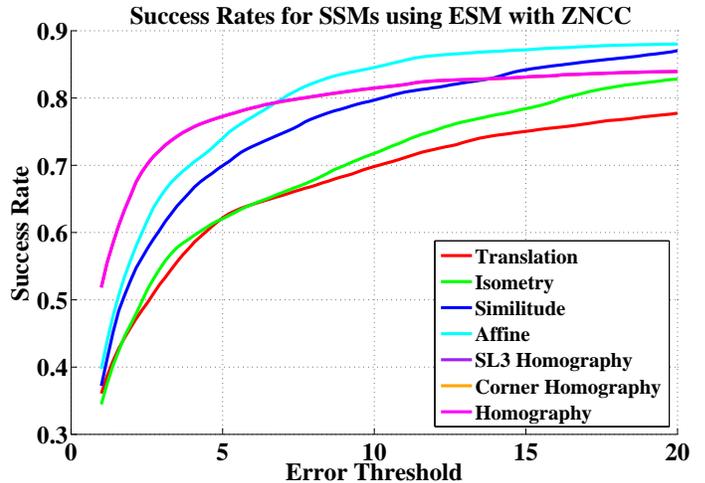}
\caption{Success Rates for all SSMs using ESM with ZNCC. Note that homography has 3 parameterizations that overlap perfectly. These plots were generated using corresponding low DOF ground truth for each SSM.} 
\label{fig_ssm}
\end{center}
\end{figure}

To conclude the analysis in this section, we tested the performance of different SSMs against each other and the results are reported in Fig. \ref{fig_ssm} using ESM with ZNCC. The plots for each SSM were generated by using the corresponding low DOF ground truth.
As stated before, we were expecting lower DOF trackers to perform better here but this is not the case since higher DOF trackers seem to perform better with the exception of affine which is better than homography for larger values of $ t_p $. However, the increased robustness of low DOF SSMs is at least partially apparent in the fact that their curves approach those of homography as $ t_p $ increases with several surpassing it too. Thus, though they may not be as precise as homography, they do tend to be more resistant to complete drift.
In fact, a general trend noticeable from the SR plots for high DOF SSMs, not only in Fig. \ref{fig_ssm} but also others analyzed earlier, is that, unlike low DOF SSMs and learning based trackers, their SR does not continue to increase through the entire range of $ t_p $ but instead flattens out after a certain point (often for $ t_p<10 $). This results from the fact that, as long as these trackers work, they track the object very precisely but once they diverge, they do not drift off gradually but rather lose track quite abruptly.

Finally, it can be noted  that all three parameterizations of homography have exactly identical performance with their plots showing perfect overlap. This indicates that the theoretical justification given in \cite{Benhimane07_esm_journal} for parameterizing ESM with $\mathcal{SL}3$ has little practical significance. 
This, in turn, may also suggest that, contrary to the assumption in \cite{Benhimane07_esm_journal}, the reason for ESM's superior performance has more to do with its use of the information from both $ I_0 $ and $ I_t $ rather than with it providing a pseudo second order convergence (opposed to LK's first order convergence). 
\section{Conclusions and Future Work}
We formulated a novel method to decompose registration based trackers into sub modules and tested several different combinations of methods for each sub module to gain interesting insights into the strengths and weaknesses of these methods. We also obtained some rather surprising results that proved previously published theoretical analysis to be somewhat inaccurate in practice, thus demonstrating the usefulness of our framework in testing out new ideas in the domain of registration based tracking. We also make publicly available the open source modular tracking framework so all results can be reproduced. This framework, with its highly efficient and ROS compatible C++ implementations for several well established trackers, will hopefully address practical tracking needs of the wider robotics community too. 

{\small
\bibliographystyle{ieee}
\bibliography{E:/UofA/Thesis/References/references}
}

\end{document}